%% file: main.tex
\documentclass[conference]{IEEEtran}
\IEEEoverridecommandlockouts
\usepackage{cite}
\usepackage{amsmath,amssymb,amsfonts}
\usepackage{algorithmic}
\usepackage{graphicx}
\usepackage{textcomp}
\usepackage{xcolor}
\usepackage{adjustbox}
\usepackage{xspace}
\usepackage{multirow}
\usepackage{authblk}
\usepackage[pagebackref=true,breaklinks=true,colorlinks,bookmarks=false]{hyperref}

\makeatletter
\DeclareRobustCommand\onedot{\futurelet\@let@token\@onedot}
\def\@onedot{\ifx\@let@token.\else.\null\fi\xspace}

\def\etal{\emph{et al}\onedot}
\makeatother

\def\BibTeX{{\rm B\kern-.05em{\sc i\kern-.025em b}\kern-.08em
    T\kern-.1667em\lower.7ex\hbox{E}\kern-.125emX}}
\begin{document}

\title{UWFormer: Underwater Image Enhancement via a Semi-Supervised Multi-Scale Transformer
% {\footnotesize \textsuperscript{*}Note: Sub-titles are not captured in Xplore and
% should not be used}
% \thanks{Identify applicable funding agency here. If none, delete this.}
}

\author[1\IEEEauthorrefmark{1}]{Weiwen Chen}
\author[1\IEEEauthorrefmark{1}]{Yingtie Lei\thanks{\IEEEauthorrefmark{1} Equal contributions}}
\author[1]{Shenghong Luo}
\author[2]{Ziyang Zhou}
\author[2]{Mingxian Li}
\author[1\IEEEauthorrefmark{2}]{Chi-Man Pun\thanks{\IEEEauthorrefmark{2} Corresponding author}}
\affil[1]{University of Macau, Macau, Macao}
\affil[2]{Huizhou University, Huizhou, China}

\maketitle

\begin{abstract}
Underwater images often exhibit poor quality, distorted color balance and low contrast due to the complex and intricate interplay of light, water, and objects. Despite the significant contributions of previous underwater enhancement techniques, there exist several problems that demand further improvement: (i) The current deep learning methods rely on Convolutional Neural Networks (CNNs) that lack the multi-scale enhancement, and global perception field is also limited. (ii) The scarcity of paired real-world underwater datasets poses a significant challenge, and the utilization of synthetic image pairs could lead to overfitting. To address the aforementioned problems, this paper introduces a Multi-scale Transformer-based Network called UWFormer for enhancing images at multiple frequencies via semi-supervised learning, in which we propose a Nonlinear Frequency-aware Attention mechanism and a Multi-Scale Fusion Feed-forward Network for low-frequency enhancement. Besides, we introduce a special underwater semi-supervised training strategy, where we propose a Subaqueous Perceptual Loss function to generate reliable pseudo labels. Experiments using full-reference and non-reference underwater benchmarks demonstrate that our method outperforms state-of-the-art methods in terms of both quantity and visual quality. The code is available at https://github.com/leiyingtie/UWFormer.
\end{abstract}


\section{Introduction}
Underwater imaging is vital for the study of oceanic ecosystems, surveying marine resources and monitoring the environment. However, the quality of underwater images is often poor due to a combination of factors notably light scattering due to particle interactions, color cast caused by light wavelength absorption, turbidity from suspended particles, the influence of water depth and temperature, variations in light penetration, motion-induced blurring, and surface disturbances. As a result, underwater images often exhibit reduced clarity, distorted color balance and low contrast. To mitigate these challenges, various image enhancement techniques have been proposed. These techniques fall broadly into two categories: traditional methods and learning-based solutions~\cite{yang2019depth}. Traditional underwater image enhancement methods often rely on mathematical models of the underwater environment and prior knowledge such as the medium transmission and the dark channel prior~\cite{drews2016underwater,li2016underwater,ding2021depth,hou2020underwater,akkaynak2019sea}. Additionally, they often require manual input of predefined parameters, which limits their adaptability to diverse and dynamic underwater environments, as depicted in Figure~\ref{fig:teaser} (b).

\input{figtex/1_teaser}

% In contrast, deep learning methods utilize neural networks to automatically learn features for image enhancement. Previous works have demonstrated that methods such as~\cite{fabbri2018enhancing, yu2019underwater, liu2019underwater, islam2020fast, chen2023large} have better performance compared to traditional physical modeling methods. These methods can be trained using artificially synthesized underwater images, eliminating the need for laborious manual operations and achieving state-of-the-art performance in extracting image information without prior knowledge of the environment. Despite the advantages of learning-based methods for enhancing underwater images, most of these methods rely on supervised learning and require paired datasets, which can limit their generalization performance, as shown in Figure~\ref{fig:teaser} (c, d). Moreover, training on synthesized underwater images~\cite{shrivastava2017learning, reed2016generative} is limited in accurately reflecting real underwater scenes. 

Learning-based techniques and data-driven approaches seek to overcome these limitations by learning directly from data. Previous works~\cite{fabbri2018enhancing, yu2019underwater, liu2019underwater, islam2020fast, li2023large, docshadow_sd7k} have demonstrated that improved performance can be achieved using deep learning and multi-frequency methods. However, most existing methods still rely on supervised learning, which requires large paired datasets for the model's generalization. Although training on synthetically generated underwater images ~\cite{shrivastava2017learning,reed2016generative} can be beneficial, it exhibits limitations in effectively capturing the intricate characteristics of real-world underwater scenes. Given the limited availability of large paired datasets for underwater image enhancement, it becomes imperative to harness the potential of features within unpaired datasets.

In addition, existing deep learning-based underwater enhancement techniques still rely on CNNs. Although CNNs excel at capturing local patterns and features within images through their convolutional and pooling layers, they might struggle with capturing long-range dependencies and global context. The emergence of Vision Transformers~\cite{dosovitskiy2020vit} allows for the capture of both local and global information. This capacity for cross-positional interaction grants Vision Transformers an advantage in handling tasks requiring broader context and intricate relationships within images. Employing coarse-to-fine techniques constitutes a viable strategy for optimizing image processing efficiency and mitigating computational resource requirements.
Another notable limitation of these prior works is the lack of multi-scale enhancement, which is particularly important as water interacts with light in a way that varies depending on the wavelengths of light. This causes different color spectra to undergo distinct attenuation rates during their passage through the water. Consequently, adopting a multi-scale methodology that optimizes for multiple wavelengths stands as a more precise strategy for encapsulating the intricate interplay of light, water, and objects. Such an approach has the potential to yield superior image enhancement.
To address the aforementioned limitations, this paper introduces the following contributions:

\begin{enumerate}
    \item We propose UWFormer, a novel semi-supervised Multi-scale Transformer for enhancing low-frequency features. UWFormer consists of two novel modules for underwater restoration: the Nonlinear Frequency-aware Attention module and the Multi-scale Fusion Feed-forward network. The incorporation of these two modules enables frequency-aware attention and varying receptive fields, which are shown to significantly improve the restoration performance.
    
    \item We propose a unique semi-supervised framework for underwater settings. A Subaqueous Perceptual Loss function is used for generating positive pseudo labels. This approach allows the teacher's output to provide reliable pseudo ground truth for the student model.
    
    \item Extensive experiments on six benchmark datasets demonstrate that our UWFormer outperforms current state-of-the-art methods in terms of visual quality across various underwater scenarios.
    
\end{enumerate}

\input{figtex/semi.tex}
\section{Related Work}
\subsection{Traditional Underwater Image Enhancement}
Most of the traditional underwater image enhancement solutions are physics-based~\cite{sahu2014survey}, such as Ancuti \etal~\cite{ancuti2012enhancing} color correction and contrast enhancement method, rely on weight mapping and image combination techniques for the final output. Additionally, MMLE~\cite{9788535} utilizes locally adaptive contrast enhancement and color loss minimization to increase image clarity and enhance contrast. GDCP~\cite{Yan2018Generalization}proposed the generalized dark channel prior approach to restoring underwater images. Over the past few years, learning-based methods have emerged as powerful underwater image enhancement techniques.

\subsection{Learning-based Underwater Image Enhancement}
Recent years have seen rapid development of learning-based methods. Ucolor~\cite{li2021underwater} uses a medium transmission-guided multicolor spatial embedding network to enhance visual quality effectively. Shallow-UWNet~\cite{2021Shallow} is a lightweight neural network structure that enhances underwater images with fewer parameters. UWCNN~\cite{2019Underwater} proposes an underwater image enhancement CNN model based on underwater scenes prior. PWRNet~\cite{huo2021efficient} utilizes a wavelet boost learning strategy to progressively enhance underwater images in both spatial and frequency domains. FuineGAN~\cite{islam2020fast} enhances image brightness, contrast, and color balance in real-time, facilitating underwater target detection and identification. TOPAL~\cite{TOPAL} proposes a fusion network that adversarially and perceptually orients the target. UGAN ~\cite{fabbri2018enhancing} aims to enhance the performance of Autonomous Underwater Vehicles in vision-driven tasks.

% \ke{need some sentences to describe the difference between this work and previous works}
% \subsection{Semi-supervised Learning}
% Semi-supervised learning methods, including Mean Teacher~\cite{tarvainen2017mean}, Virtual Adversarial Learning~\cite{miyato2018virtual}, and FixMatch~\cite{sohn2020fixmatch}, have proliferated in recent years. However, there is minimal research on using semi-supervised Transformers for enhancing underwater images. Semi-supervised technique exploits the small amount of labeled data and large amount of unlabeled data to train the Transformer model. As a result, the model's accuracy and generalization capabilities increase, and it eliminates the need for large paired image datasets required for learner-based methods. SemiFormer~\cite{weng2022semi}, a semi-supervised framework that includes a Transformer stream, convolutional stream, and meticulously planned fusion module, is a noteworthy contribution to this field. Additionally, Semi-VIT~\cite{cai2022semi} features a large Transformer model with a greater number of layers and channels to increase accuracy.

% \input{figtex/2_model}

\input{figtex/SMViT.tex}

\section{Methodology}
\subsection{Overview}
The overall architecture of the proposed UW-Former is illustrated in Figure~\ref{fig:stmnet}.  To enable multi-scale enhancement and save computational resources, we adopt lossless downsampling and reconstruction. We utilize the Discrete Wavelet Transform (DWT) and its inverse (IDWT)~\cite{mallat1989theory,li2022wavenhancer} to enhance individual frequencies using distinct methods. Since high-frequency components, such as texture and edges, often undergo minor alteration~\cite{liang2021high}, we use a simple residual network composed of Fast Fourier Convolutions to enhance these frequencies. Conversely, we introduce our proposed Transformer architecture to refine low-frequency components, which contain essential color information. This allows for fewer computing resources to be allocated to the high-frequency domain with lower variability, while dedicating more resources to the domain that requires more learning.

\subsection{Transformer Architecture}
The Vision Transformer model can effectively utilize self-attention to capture both global and local information concurrently, enabling the extraction of global features~\cite{zhou2024qean,chen2023medprompt}. This characteristic can be particularly advantageous for enhancing underwater images. Furthermore, coarse-to-fine strategies have been demonstrated to be effective for image enhancement~\cite{Li_2023_ICCV,ijcai2023p129,luo2023devignet} and other tasks~\cite{liu2023explicit,Zuo2023BrainFN,liu2023coordfill,liu2024depth,li2023cee,li2022monocular,jing2024estimating,jing2023ta}. Accordingly, we propose our Multi-scale Transformer, which stacks Transformers with multi-scale inputs and progressively refines image quality from bottom to top using the coarse-to-fine approach. This multi-scale architecture allows information in the image to flow through the network while tightly integrating low-level and high-level information into a UNet-like structure. The whole process is depicted in Figure~\ref{fig:stmnet}.

More specifically, the UWFormer model contains four blocks for the feature extraction process at the start and four blocks for the feature reconstruction process at the end of the network. The model consists of a UNet architecture that has been optimized for multi-scale structures, with each block comprised of a Nonlinear Frequency-Aware Attention (NFA) layer and a Multi-Scale Fusion Feed-forward Network (MSFN). In the middle of the network are four encoders and three decoders, with a progressive increase in the number of blocks in each encoder and a progressive decrease in the number of blocks in each decoder. At each scale, we input $X_1 \in \mathbb{R}^{C \times H/2 \times W/2}$, $X_2 \in \mathbb{R}^{C \times H/4 \times W/4}$ and $X_3 \in \mathbb{R}^{C \times H/8 \times W/8}$ in addition to the original input image, $X \in \mathbb{R}^{C \times H \times W}$. Before being fed into each scale's encoder, the images undergo feature extraction using a convolution-based process and are then concatenated with the output of the previous encoder.

% We introduce the Asymmetric Feature Fusion (AFF) method as portrayed in Figure~\ref{fig:aff}(b) to promote the amalgamation of information across various levels~\cite{cho2021rethinking}. AFF concatenates four diverse features and then restores the channel dimension. This is accomplished using a $1 \times 1$ and a $3 \times 3$ convolution. The resultant feature map is then combined with the corresponding decoder module using skip connections. 

% For instance, at level one, this process can be represented as Eq. \ref{eq:aff}.
% \begin{equation}
%     \operatorname{AFF}_{1}^{\text {out }}=\mathrm{AFF}_{1}\left(\mathrm{~EC}_{1}^{\text {out }},\left(\mathrm{EC}_{2}^{\text {out }}\right)^{\uparrow},\left(\mathrm{EC}_{3}^{\text {out }}\right)^{\uparrow},\left(\mathrm{EC}_{4}^{\text {out }}\right)^{\uparrow}\right)
%     \label{eq:aff}
% \end{equation}

% Here, $\mathrm{~EC}_{*}^{\text {out }}$ represents the output of each encoder, $\uparrow$ is used for upsampling, and $\downarrow$ denotes downsampling.

% \input{tables/userstudy}

\subsubsection{Nonlinear Frequency-Aware Attention (NFA)}
% Existing Vision Transformer (ViT) backbones often employ downsampling techniques to reduce the computational costs of self-attention for high-resolution inputs~\cite{wang2022pvt}. Nonetheless, common downsampling methods like average pooling result in irreversible information loss. To address this drawback, researchers like Yao \etal~\cite{yao2022wave} have proposed using DWT and IDWT to downsample and reconstruct feature maps. However, relying on linear layers for learning image information may lead to the inability of the model to capture the non-linear patterns present in images, potentially leading to issues like overfitting and the presence of excessive parameters.

Researchers like~\cite{yao2022wave,Chen2024-dg,huang2023mr,10365931} have proposed to downsample and reconstruct feature maps in attention mechanism. However, relying on linear layers for learning image information may lead to the inability of the model to capture the non-linear patterns present in images, potentially leading to issues like overfitting and the presence of excessive parameters. To address this issue, we propose a new module called Nonlinear Frequency-aware Attention. In this module, we replace the conventional linear layers with simple convolutional layers and apply DWT and IDWT to enable lossless downsampling and reconstruction of the feature map. This results in a significant improvement of multi-scale enhancement for underwater scenarios and enhances the non-linear expressive power of the attention mechanism.

NFA processes the feature map $X \in \mathbb{R}^{C \times H \times W}$ through two pathways. The first pathway generates the Query $Q \in \mathbb{R}^{4C \times H \times W}$ by passing $X$ through a $1 \times 1$ and $3 \times 3$ Depth-wise convolutional (DWConv) layer. In the second pathway, we concatenate the result of the DWT of $X$ into four frequency domains and apply them to two sets of $1 \times 1$ and $3 \times 3$ convolutional layers to produce the Key $K \in \mathbb{R}^{16C \times H/2 \times W/2}$ and Value $V \in \mathbb{R}^{16C \times H/2 \times W/2}$. IDWT is also performed on the locally concatenated contextualized downsampled feature $X_c$ as a byproduct, which, according to wavelet theory, preserves every detail of the original input $X$. The Query, Key, and Value are subsequently reshaped into $\hat{Q} \in \mathbb{R}^{4C \times HW}$, $\hat{K} \in \mathbb{R}^{HW \times 4C}$, and $\hat{V} \in \mathbb{R}^{HW \times 4C}$, respectively, for the purpose of matrix multiplication. By implementing this process of DWT-Convolution-IDWT in the NFA, a more robust local contextualization with an enlarged receptive field is achieved. 

Finally, we formulate the NFA as Eq.\ref{eq:nfa}:
\begin{equation}
    \begin{aligned}
        &\operatorname{NFA}=\operatorname{MultiHead}\left(Q, K, V, X^{r}\right),\\
        &\operatorname{MultiHead}\left(Q, K, V, X^{r}\right)=\left[head_0, ..., head_j, X^{r}\right]W^O,\\
        &head_{j}=\operatorname{Attention}{\left(Q_{j}, K_{j}, V_{j}\right)}=\operatorname{Softmax}\left(\frac{Q_{j} K_{j}^{T}}{\sqrt{D_{h}}}\right) V_{j},
    \end{aligned}
    \label{eq:nfa}
\end{equation}
where $\left[ \cdot \right]$ stands for concatenation and $W^O$ is the transformation matrix.

\subsubsection{Multi-Scale Fusion Feed-forward Network (MSFN)}

Inspired by FFC, which uses two channels for image processing, we propose the Multi-Scale Fusion Feed-forward Network. Similar to FFC, MSFN is a multi-scale structure comprising two connected pathways - a local pathway performing regular convolutions on specific input feature channels, and a global pathway, which operates larger convolutions in the global domain. Each path acquires diverse knowledge that is mutually complementary, allowing features to be improved at various scales across the network, facilitated by the exchange of information between them. Our design allows us to enhance features at multiple scales, a key benefit of MSFN over other architectures.

The MSFN model is designed to extract both detailed and global information from an underwater feature map $X \in \mathbb{R}^{C \times H \times W}$. Specifically, the local branch of MSFN employs regular $3 \times 3$ depth-wise convolution to extract detailed information, resulting in feature maps $X_{ll}$ and $X_{lg}$ with the same dimensions as $X$. On the other hand, the global branch of MSFN utilizes a larger $5 \times 5$ convolution kernel to capture more global information, resulting in feature maps $X_g$ and $X_{gl}$ of the same size as $X_l$ and $X_{lg}$. To integrate the local and global information, the feature maps from both branches are combined in a cross-scale manner. 

This process can be represented mathematically as shown in Equation \ref{eq:msfn}, which yields a superior enhancement quality:
\begin{equation}
    \begin{aligned}
        &Y = f_{1x1}\left[\sigma(f_{3  \times 3}^{dwc}(X_l)), \sigma(f_{3 \times 3}^{dwc}(X_g))\right],\\
        & X_l = \left[ \sigma(f_{3 \times 3}^{dwc}(X_{ll})), \sigma(f_{5 \times 5}^{dwc}(X_{gl}))\right],\\
         & X_g = \left[ \sigma(f_{3 \times 3}^{dwc}(X_{lg})), \sigma(f_{5 \times 5}^{dwc}(X_{gg}))\right],
    \end{aligned}
    \label{eq:msfn}
\end{equation}
where $\sigma (\cdot)$ indicates GeLU activation, $f_{1 \times 1}$ is $1 \times 1$ convolution, $f_{3 \times 3}^{dwc}$ and $f_{5 \times 5}^{dwc}$ denote  $3 \times 3$ and $5 \times 5 $ depth-wise convolution, $\left[ \cdot \right]$ means concatenation.

\input{figtex/results_ref.tex}

\subsection{Subaqueous Perceptual Loss}
The limited availability of labeled underwater enhancement datasets necessitates the integration of unlabeled data to improve performance. Generally speaking, semi-supervised strategies are suitable for this scenario, but common semi-supervised strategies have not been optimized specifically for underwater augmentation, which may result in ultimately unsatisfactory outcomes. To this end, we introduce a new Subaqueous Perceptual Loss in a special-designed semi-supervised training strategy to learn from both labeled and unlabeled data. 

The Subaqueous Perceptual Loss is an underwater quality loss function defined in the CIELab~\cite{yang2015underwater, backhaus2011color} color space that enables the teacher model to produce high-quality reliable pseudo labels, which can be written as Equation~\ref{eq:spl}:

\begin{equation}
    SPL=c_{1} \times \sigma_{c}+c_{2} \times \operatorname{c}_{l}+c_{3} \times \mu_{s},
\label{eq:spl}
\end{equation}
where $c_{1}, c_{2}, c_{3}$ are weighted coefficients, which are assigned the values 0.4680, 0.2745 and 0.2576 respectively; $\sigma_{c}$ is the chroma standard deviation; $\operatorname{c}_{l}$ is the luminance contrast; $\mu_{s}$ is the average saturation.

The proposed UWFormer comprises two networks with matching structures, referred to as the teacher model and the student model~\cite{tarvainen2017mean}. The primary distinction between the two networks lies in how their weights are updated. More specifically, we utilize normal gradient descent to update the student's weights ($\theta_{s}$), whereas the teacher's weights ($\theta_{t}$) are updated based on the exponential moving average (EMA) of the student's weights. Through SPL, the outputs of the teacher model will significantly rectify the generation results of the student model and ultimately output desirable augmented underwater images.

\input{tables/ref_com}

\input{tables/noref_com}
\section{Experiments}
\subsection{Experiment Settings}
In this section, we describe in detail the dataset, evaluation metrics and implementation details. It is worth noting that like previous works, we do not place emphasis on network speed since all the test data is $256 \times 256$, so all the methods can generate results in a very short period of time.
\subsubsection{Dataset}
The training set for our experiment is made up of 2800 labeled image pairs and 2800 unlabeled images. The labeled dataset is randomly selected at a 5:2 ratio from EUVP~\cite{islam2020fast} and UIEB~\cite{li2019underwater}. The UIEB dataset consists of 890 real-world underwater images with accompanying ground truths. The unlabeled images are obtained in a ratio of 11:17 from the unpaired data in EUVPUN~\cite{islam2020fast} and RUIE~\cite{liu2020real}. They represent a diverse range of underwater scenes, water types, and lighting conditions, thus enhancing the generalization ability of our model. 

For testing, we prepared two datasets: full-reference and no-reference. The full-reference dataset contains the remaining 90 paired images from UIEB and the remaining 200 paired images from EUVP. The no-reference dataset includes 1200 images from EUVPUN, full images from U45~\cite{li2019fusion}, the remaining 1930 images from RUIE, and all images from UIEB-60~\cite{li2019underwater}. By incorporating different underwater scenarios, water types, and lighting conditions, these datasets can provide a comprehensive assessment of the model's performance.

\input{figtex/results_noref.tex}

\subsubsection{Evaluation Metrics}
Our study employs several full-reference evaluation metrics, including Peak Signal-to-Noise Ratio (PSNR), Structural Similarity Index (SSIM), and Learned Perceptual Image Patch Similarity (LPIPS)~\cite{zhang2018unreasonable}. 

For no-reference evaluation metrics, we have selected two metrics. The first is the Underwater Image Quality Metric (UIQM)~\cite{panetta2015human}, which considers the unique degradation mechanisms and imaging characteristics of underwater images. The second metric is the Underwater Color Image Quality Evaluation (UCIQE)~\cite{yang2015underwater}, which is a linear combination of colorfulness, saturation, and contrast. These metrics are chosen for their ability to accurately assess the quality of underwater images.
\subsubsection{Implementation Details}
Our model is implemented with PyTorch and trained using the Adam optimizer with default parameters on four NVIDIA RTX 8000. A batch size of 4 is used in training for 200 epochs, and the learning rate is reduced by a factor of 0.1 at epoch 100. The model is trained in a semi-supervised strategy for 200 epochs. Data augmentation techniques included random flipping, rotation, cropping, and resizing.

\subsection{Comparisons with State-of-the-Arts}
We compared our UWFormer model with nine existing state-of-the-art methods, GDCP~\cite{Yan2018Generalization} and MMLE~\cite{9788535}, UWCNN~\cite{2019Underwater}, TOPAL~\cite{TOPAL}, UGAN~\cite{fabbri2018enhancing}, FunieGAN~\cite{islam2020fast}, Ucolor~\cite{li2021underwater}, PWRNet~\cite{huo2021efficient}, and ShallowUW~\cite{2021Shallow}. All methods have been retrained on our datasets with 200 epochs using their official implementations.

Table~\ref{tab:ref_com} presents the quantitative results of the full-reference datasets. It reveals that our UWFormer outperforms the existing methods in terms of both image similarity and visual quality in most metrics. Multi-scale Transformer architecture afforded a larger receptive field during supervised learning. This enabled superior performance over other methods in color restoration and detail preservation. The superiority of our proposed method is also evident in Figure~\ref{fig:ref}, where UWFormer achieves pleasing aesthetic outcomes, whereas the compared methods suffer from issues such as color cast and over-enhancement.

The quantitative results of the no-reference datasets are presented in Table~\ref{tab:noref_com}. Our UWFormer achieves the best performance for most metrics and ranks second in three metrics in the RUIE, and UIEB-60 datasets. Our proposed SPL function successfully achieved good performance in semi-supervised learning while maintaining supervised performance. Notably, generative models may hold an advantage for unsupervised tasks. We have provided visual comparisons in Figure~\ref{fig:noref}, which demonstrate that our proposed method produces visually satisfying results, outperforming the existing methods.

\subsection{Ablation Studies}
We carry out ablation studies to assess the effectiveness of our proposed UWFformer approach by testing its individual components, and we replace NFA and MSFN with MDTA and GDFN since they are most commonly used mechanisms in Tranformer. These components are as follows: 

\begin{enumerate}
    \item \textbf{UNet:} A baseline;
    \item \textbf{W/O SPL:}  Let teacher model generate pseudo labels without SPL;
    \item \textbf{MDTA+MSFN:} Replacing NFA with MDTA~\cite{zamir2022restormer};
    \item \textbf{NFA+GDFN:} Replacing MSFN with GDFN~\cite{zamir2022restormer};
    \item \textbf{Supervised:} The full-supervised training strategy without semi-supervision;
    \item \textbf{Ours:} Our full UWFormer architecture with the semi-supervised training strategy.
\end{enumerate}

The results of our ablation studies on full-reference and no-reference datasets are presented in Tables~\ref{tab:ref_com} and \ref{tab:noref_com}, respectively. Our full version outperforms all other variants. We note that each module makes a significant contribution to the overall performance. Notably, each component proposed in this work contributed to improved performance.

The \textbf{baseline} experiments without our modules perform the worst. The \textbf{W/O SPL} represents the effectiveness of the proposed SPL loss function. 
Our \textbf{MDTA+MSFN} and \textbf{NFA+GDFN} experiments demonstrate the effectiveness of our NFA and MSFN, respectively. 
The \textbf{Supervised} experiment performs similarly to our full-version on the paired dataset but falls short on the no-reference metrics due to the absence of unlabeled data. In conclusion, our full-version model demonstrates excellent performance on both full-reference and non-reference datasets.
% \subsection{User Study}
% After careful consideration, we selected the top-performing three methods based on our performance metrics for a user study, which is detailed in Table~\ref{tab:user}. We randomly selected 15 sets of distinct images and 22 participants rated four sets of full-reference and four sets of no-reference results based on two criteria for full-reference: (a) Authenticity and (b) Simulation Effectiveness, indicating which image comes closest to a realistic underwater image, and which image is closest to the target image, respectively. 

% Our rankings are calculated utilizing a weighted average, where lower values corresponded to higher rankings. Our results achieved the top ranking, signifying that our results are the most preferred by the participants.

% \input{tables/userstudy.tex}
\section{Conclusion}

This paper presents UWFormer, a novel multi-scale Transformer-based network designed to enhance images across multiple frequencies using a semi-supervised learning approach. The proposed UWFormer architecture incorporates an underwater-specific design that enhances performance through the integration of a Nonlinear Frequency-aware Attention mechanism and a Multi-scale Fusion Feed-forward network. Additionally, a new loss function Subaqueous Perceptual Loss is proposed to effectively guide the teacher model in generating pseudo labels. We conduct extensive experiments on full-reference and non-reference datasets to demonstrate that UWFormer outperforms existing methods both quantitatively and qualitatively. 
% In future work, we plan to further explore the incorporation of prior knowledge specific to underwater enhancement into our model, aiming to further improve its performance and robustness.

\section*{Acknowledgment}
This work was supported in part by the Huizhou Daya Bay Science and Technology Planning Project under Grant No.2020020003, in part by  the Science and Technology Development Fund, Macau SAR, under Grant 0087/2020/A2 and Grant 0141/2023/RIA2.

\bibliographystyle{IEEEtran}
\bibliography{ref}

\end{document}

%% file: figtex/1_teaser.tex
\begin{figure}[t]
    \begin{minipage}[b]{1.0\linewidth}
        \begin{minipage}[b]{.32\linewidth}
            \centering
            \centerline{\includegraphics[width=\linewidth]{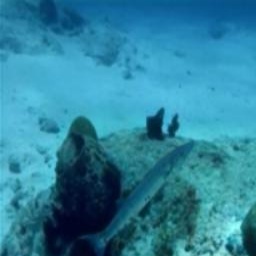}}
            \centerline{(a) Input}\medskip
        \end{minipage}
        \hfill
        \begin{minipage}[b]{.32\linewidth}
            \centering
            \centerline{\includegraphics[width=\linewidth]{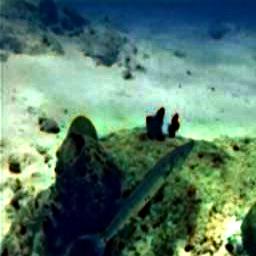}}
            \centerline{(b) GDCP}\medskip
        \end{minipage}
        \hfill
        \begin{minipage}[b]{0.32\linewidth}
            \centering
            \centerline{\includegraphics[width=\linewidth]{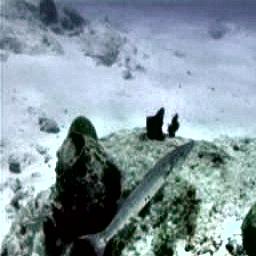}}
            \centerline{(c) MMLE}\medskip
        \end{minipage}
    \end{minipage}
    \begin{minipage}[b]{1.0\linewidth}
        \begin{minipage}[b]{.32\linewidth}
            \centering
            \centerline{\includegraphics[width=\linewidth]{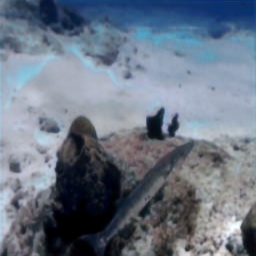}}
            \centerline{(d) Ucolor}\medskip
        \end{minipage}
        \hfill
        \begin{minipage}[b]{.32\linewidth}
            \centering
            \centerline{\includegraphics[width=\linewidth]{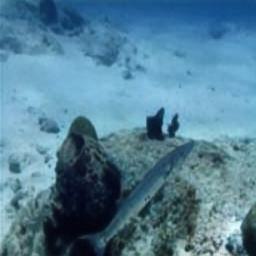}}
            \centerline{(e) Ours}\medskip
        \end{minipage}
        \hfill
        \begin{minipage}[b]{0.32\linewidth}
            \centering
            \centerline{\includegraphics[width=\linewidth]{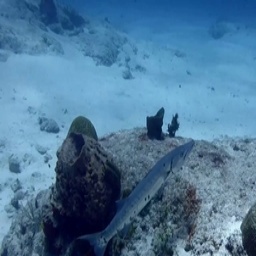}}
            \centerline{(f) Target}\medskip
        \end{minipage}
    \end{minipage}
    \caption{
    The figure illustrates a comparison between (a) underwater image  and (f) its enhanced version obtained from the EUVP dataset. Although (b),(c) traditional enhancement, and (d) deep learning models have been employed, they still manifest imperfections. By contrast, (e) our method effectively enhances the image.
    } 
    \label{fig:teaser}
\end{figure}

%% file: figtex/semi.tex
\begin{figure*}[ht]
    \centering
    \resizebox{0.95 \linewidth}{!}{
    \includegraphics[width=\linewidth]{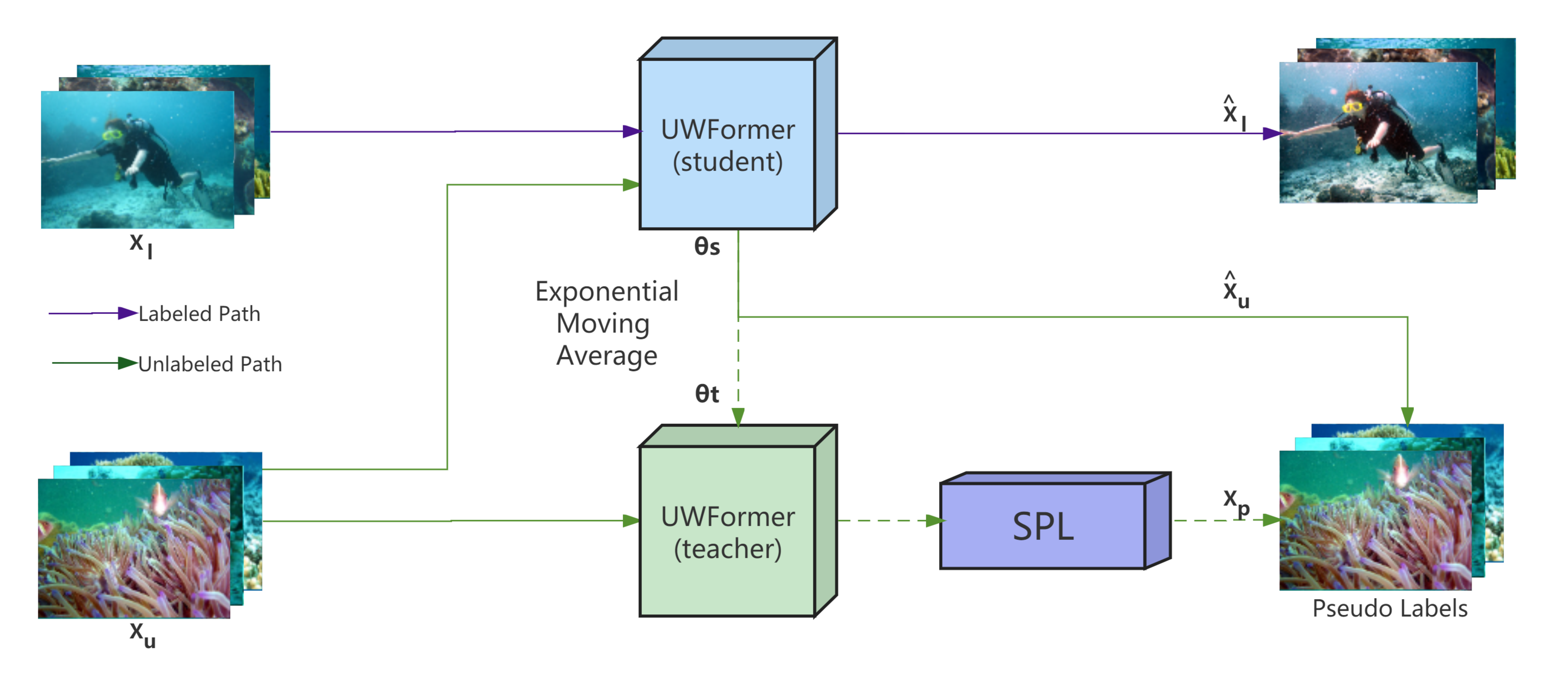}
    }
    \caption{
     Proposed underwater semi-supervised training strategy.% of ours. 
    Labeled images $X_l \in \mathbb{R}^{C \times H \times W}$ and unlabeled images $X_u \in \mathbb{R}^{C \times H \times W}$ are fed to: (i) the student model for training, and (ii) the teacher model for unlabeled prediction. The student model outputs labeled results $\hat{X_l}$ and unlabeled results $\hat{X_u}$, with the $\hat{X_l}$ converging using a supervised loss function. Note that the outputs of the teacher model are evaluated and updated in real-time using the proposed Subaqueous Perceptual Loss (SPL) to generate pseudo labels $X_p$. Finally, the unlabeled outputs of the student model $\hat{X_u}$ are constrained by $X_p$ using an unsupervised loss function. In this case, the weights of the student model are determined by both labeled data and  unlabeled data.
    }
    \label{fig:semi}
\end{figure*}

%% file: figtex/SMViT.tex
\begin{figure*}[ht]
    \begin{minipage}[b]{1.0\linewidth}
        \includegraphics[width=\linewidth]{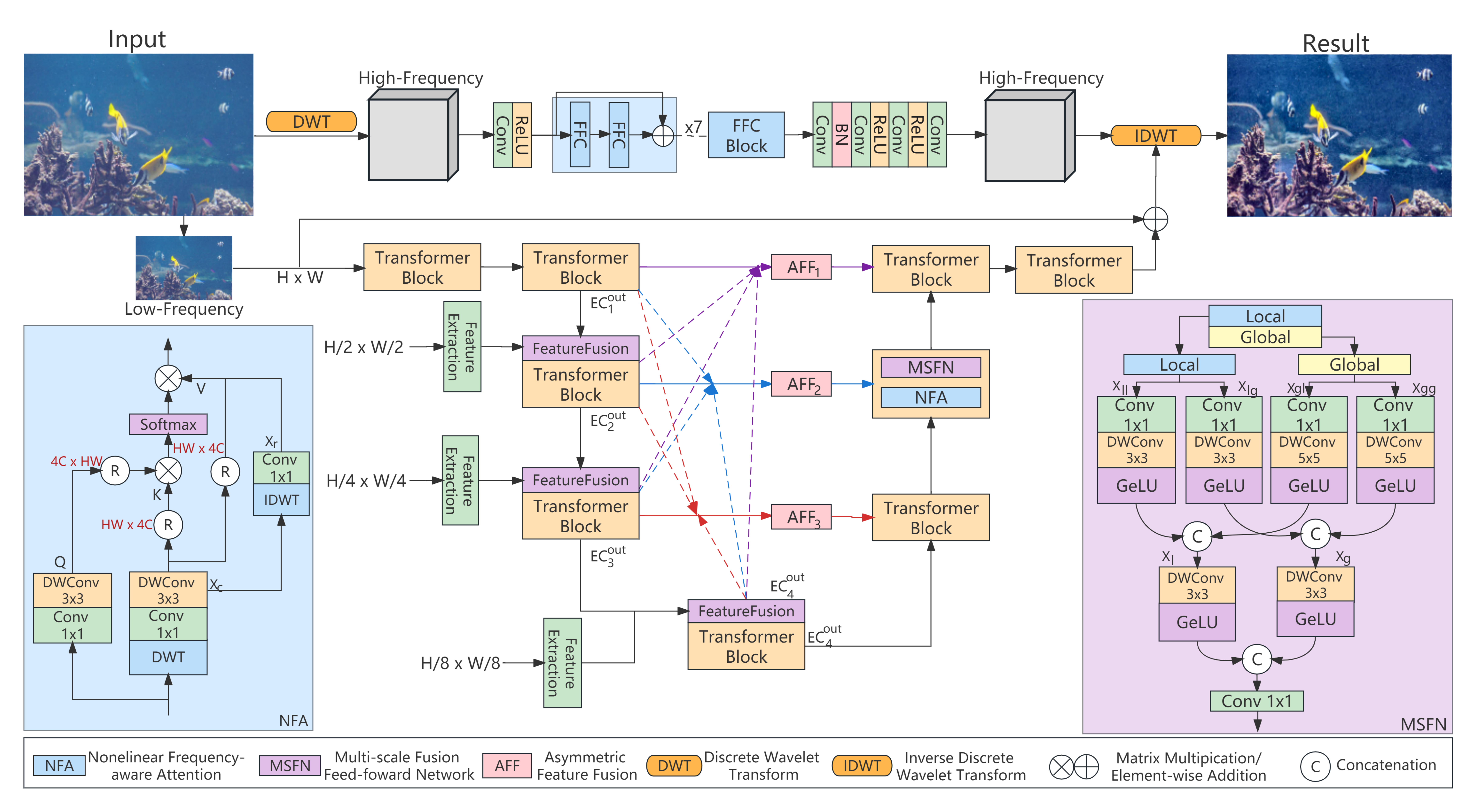}
    \end{minipage}
    \caption{
    Overall architecture of our proposed UWFormer. An input image $X \in \mathbb{R}^{C \times H \times W}$, it is first processed by DWT to produce three high-frequency images $\left\{ X_{LH}, X_{HL}, X_{HH} \right\} \in \mathbb{R}^{C \times H \times W}$ and one low-frequency image $X_{LL} \in R^{C \times H \times W}$. The three high-frequency images $X_{LH}, X_{HL}, X_{HH}$ are then merged into a 9-channel image $X_{H}$ and fed into a simple ResNet composed of FFC. The low-frequency part $X_{LL}$ is fed into our MSFormer. Note that the $X_{LL}$ at different scales are subjected to feature extraction and then fed into the MSFormer in a top-down way, where they are fused with the output of the previous encoder. Eventually, the output of the MSFormer $\hat{X_{LL}}$ and the optimized image $\hat{X_H}$ are subjected to IDWT to obtain the final result.
    }
    \label{fig:stmnet}
\end{figure*}

%% file: figtex/results_ref.tex
\begin{figure*}[!ht]

    \begin{minipage}[b]{1.0\linewidth}
        \begin{minipage}[b]{0.105\linewidth}
            \centering
            \centerline{\includegraphics[height=2.2cm,width=\linewidth]{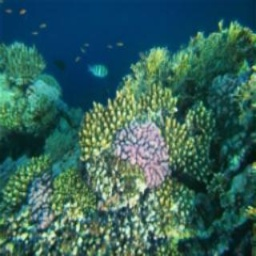}}
        \end{minipage}
        \hfill
        \begin{minipage}[b]{0.105\linewidth}
            \centering
            \centerline{\includegraphics[height=2.2cm,width=\linewidth]{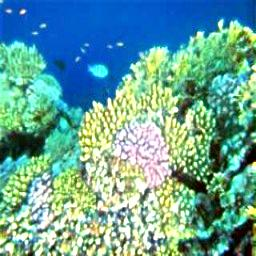}}
        \end{minipage}
        \hfill
        \begin{minipage}[b]{0.105\linewidth}
            \centering
            \centerline{\includegraphics[height=2.2cm,width=\linewidth]{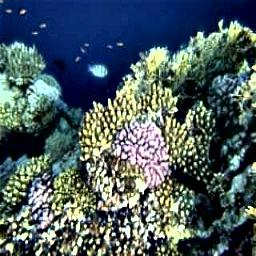}}
        \end{minipage}
        \hfill
        \begin{minipage}[b]{0.105\linewidth}
            \centering
            \centerline{\includegraphics[height=2.2cm,width=\linewidth]{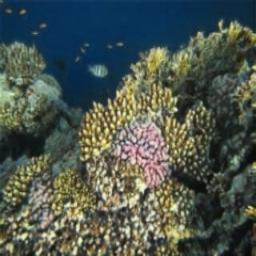}}
        \end{minipage} 
        \hfill
        \begin{minipage}[b]{0.105\linewidth}
            \centering
            \centerline{\includegraphics[height=2.2cm,width=\linewidth]{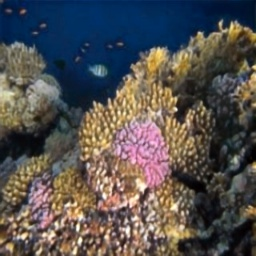}}
        \end{minipage}  
        \hfill
        \begin{minipage}[b]{0.105\linewidth}
            \centering
            \centerline{\includegraphics[height=2.2cm,width=\linewidth]{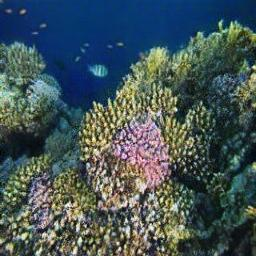}}
        \end{minipage}  
        \hfill
        \begin{minipage}[b]{0.105\linewidth}
            \centering
            \centerline{\includegraphics[height=2.2cm,width=\linewidth]{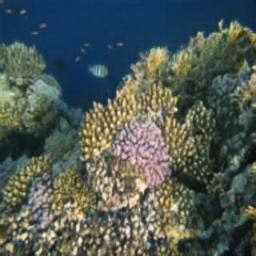}}
        \end{minipage}
        \hfill
        \begin{minipage}[b]{0.105\linewidth}
            \centering
            \centerline{\includegraphics[height=2.2cm,width=\linewidth]{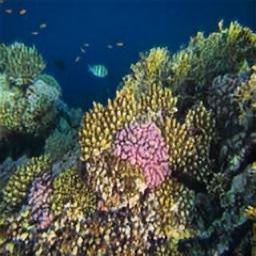}}
        \end{minipage}    
        \hfill
        \begin{minipage}[b]{0.105\linewidth}
            \centering
            \centerline{\includegraphics[height=2.2cm,width=\linewidth]{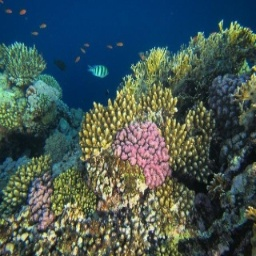}}
        \end{minipage}
    \end{minipage}

    \begin{minipage}[b]{1.0\linewidth}  
        \begin{minipage}[b]{0.105\linewidth}
            \centering
            \centerline{\includegraphics[height=2.2cm,width=\linewidth]
            {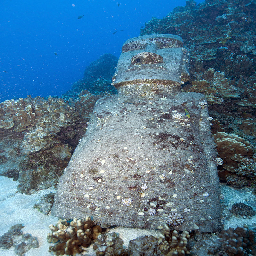}}
        \end{minipage}
        \hfill
        \begin{minipage}[b]{0.105\linewidth}
            \centering
            \centerline{\includegraphics[height=2.2cm,width=\linewidth]
            {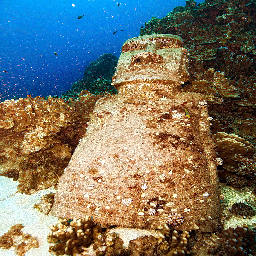}}
        \end{minipage}
        \hfill
        \begin{minipage}[b]{0.105\linewidth}
            \centering
            \centerline{\includegraphics[height=2.2cm,width=\linewidth]{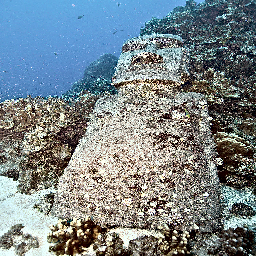}}
        \end{minipage}
        \hfill
        \begin{minipage}[b]{0.105\linewidth}
            \centering
            \centerline{\includegraphics[height=2.2cm,width=\linewidth]{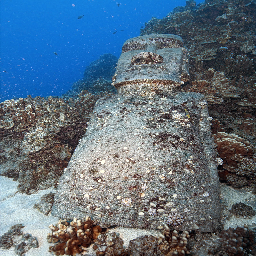}}
        \end{minipage}
        \hfill
        \begin{minipage}[b]{0.105\linewidth}
            \centering
            \centerline{\includegraphics[height=2.2cm,width=\linewidth]
            {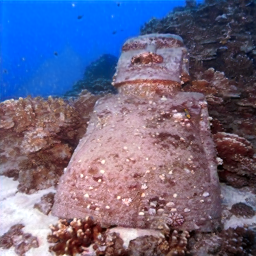}}
        \end{minipage}
        \hfill
        \begin{minipage}[b]{0.105\linewidth}
            \centering
            \centerline{\includegraphics[height=2.2cm,width=\linewidth]{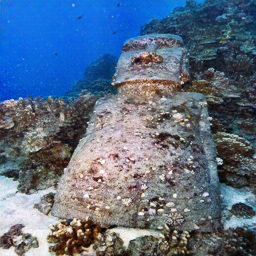}}
        \end{minipage}
        \hfill
        \begin{minipage}[b]{0.105\linewidth}
            \centering
            \centerline{\includegraphics[height=2.2cm,width=\linewidth]{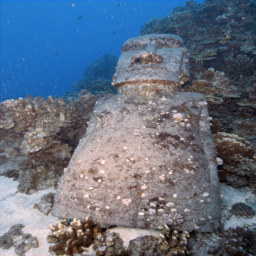}}
        \end{minipage}
        \hfill
        \begin{minipage}[b]{0.105\linewidth}
            \centering
            \centerline{\includegraphics[height=2.2cm,width=\linewidth]{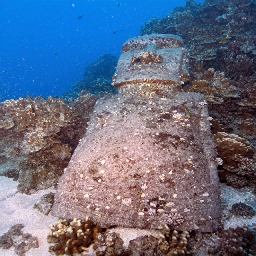}}
        \end{minipage}
        \hfill
        \begin{minipage}[b]{0.105\linewidth}
            \centering
            \centerline{\includegraphics[height=2.2cm,width=\linewidth]
            {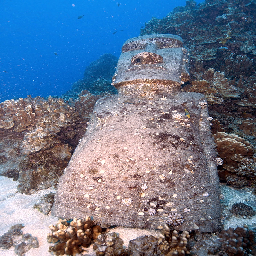}}
        \end{minipage}
    \end{minipage}

    \begin{minipage}[b]{1.0\linewidth}
        \begin{minipage}[b]{0.105\linewidth}
            \centering
            \centerline{\includegraphics[height=2.2cm,width=\linewidth]{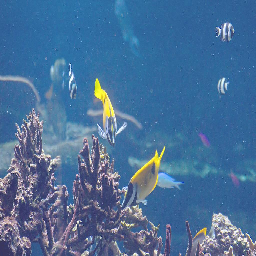}}
            \centerline{(a) Input}\medskip
        \end{minipage}
        \hfill
        \begin{minipage}[b]{0.105\linewidth}
            \centering
            \centerline{\includegraphics[height=2.2cm,width=\linewidth]{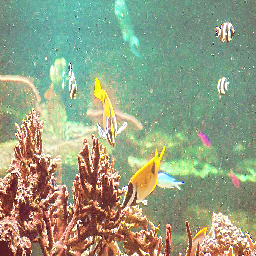}}
            \centerline{(b) GDCP}\medskip
        \end{minipage}  
        \hfill
        \begin{minipage}[b]{0.105\linewidth}
            \centering
            \centerline{\includegraphics[height=2.2cm,width=\linewidth]{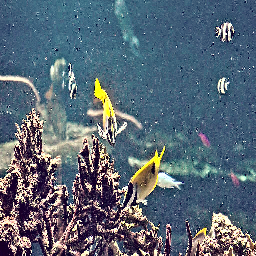}}
            \centerline{(c) MMLE}\medskip
        \end{minipage}   
        \hfill
        \begin{minipage}[b]{0.105\linewidth}
            \centering
            \centerline{\includegraphics[height=2.2cm,width=\linewidth]{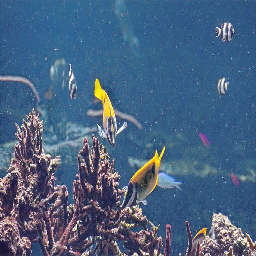}}
            \centerline{(d) UWCNN}\medskip
        \end{minipage}
        \hfill
        \begin{minipage}[b]{0.105\linewidth}
            \centering
            \centerline{\includegraphics[height=2.2cm,width=\linewidth]{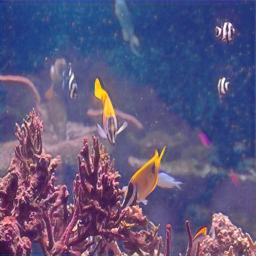}}
            \centerline{(e) Ucolor}\medskip
        \end{minipage}
        \hfill
        \begin{minipage}[b]{0.105\linewidth}
            \centering
            \centerline{\includegraphics[height=2.2cm,width=\linewidth]{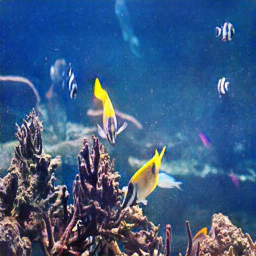}}
             \centerline{(f) Funie}\medskip
        \end{minipage}
        \hfill
        \begin{minipage}[b]{0.105\linewidth}
            \centering
            \centerline{\includegraphics[height=2.2cm,width=\linewidth]{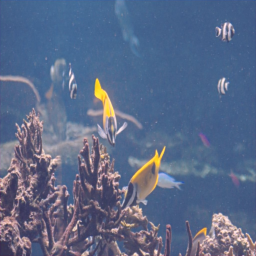}}
            \centerline{(g) ShallowUW}\medskip
        \end{minipage}
        \hfill
        \begin{minipage}[b]{0.105\linewidth}
            \centering
            \centerline{\includegraphics[height=2.2cm,width=\linewidth]{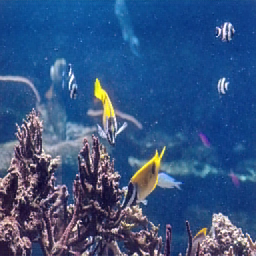}}
            \centerline{(h) Ours}\medskip
        \end{minipage}
        \hfill
        \begin{minipage}[b]{0.105\linewidth}
            \centering
            \centerline{\includegraphics[height=2.2cm,width=\linewidth]{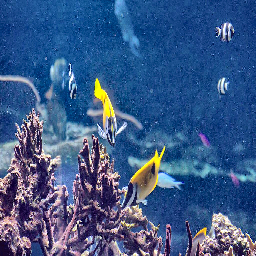}}
            \centerline{(i) Target}\medskip
        \end{minipage}
    \end{minipage}
    
    \caption{
    A visual comparison of different image enhancement methods. Methods (b) and (c) show varying degrees of over-optimization, whereas methods (d) and (g) exhibit different degrees of under-optimization. Method (f) shows relatively satisfactory optimization results, yet some optimization defects can still be observed, such as the appearance of white blocks in the middle of the last row. However, our results (h) show superior visual results in terms of color tone and details, and are the closest to the target. The first two rows of images are sourced from EUVP, while the latter two rows are obtained from UIEB.
    }
    \label{fig:ref}
\end{figure*}

%% file: tables/ref_com.tex
% Please add the following required packages to your document preamble:
% \usepackage{multirow}
% \usepackage[table,xcdraw]{xcolor}
% If you use beamer only pass "xcolor=table" option, documentclass[xcolor=table]{beamer}
\begin{table*}[ht]
\centering
\caption{Quantitative results of comparisons with the state-of-the-art on full-reference datasets. The upper half of the table is the comparison experiment, and the lower half shows the ablation experiment. The best performance is marked in bold.}
\begin{tabular}{l|cccc|cccc}
\hline
{\color[HTML]{000000} }                         & \multicolumn{4}{c|}{{\color[HTML]{000000} UIEB}}                                                                                                                                                                                 & \multicolumn{4}{c}{{\color[HTML]{000000} EUVP}}                                                                                                                                                                                 \\ \cline{2-9} 
\multirow{-2}{*}{{\color[HTML]{000000} Method}} & \multicolumn{1}{c|}{{\color[HTML]{000000} PSNR$\uparrow$}} & \multicolumn{1}{c|}{{\color[HTML]{000000} SSIM$\uparrow$}} & \multicolumn{1}{c|}{{\color[HTML]{000000} LPIPS$\downarrow$}} & {\color[HTML]{000000} UCIQE$\uparrow$} & \multicolumn{1}{c|}{{\color[HTML]{000000} PSNR$\uparrow$}} & \multicolumn{1}{c|}{{\color[HTML]{000000} SSIM$\uparrow$}} & \multicolumn{1}{c|}{{\color[HTML]{000000} LPIPS$\downarrow$}} & {\color[HTML]{000000} UCIQE$\uparrow$} \\ \hline
{\color[HTML]{000000} GDCP}                     & {\color[HTML]{000000} 14.72}                               & {\color[HTML]{000000} 0.745}                               & {\color[HTML]{000000} 0.213}                                  & {\color[HTML]{000000} 0.443}           & {\color[HTML]{000000} 13.10}                               & {\color[HTML]{000000} 0.638}                               & {\color[HTML]{000000} 0.329}                                  & {\color[HTML]{000000} 0.427}           \\
{\color[HTML]{000000} MMLE}                     & {\color[HTML]{000000} 18.24}                               & {\color[HTML]{000000} 0.767}                               & {\color[HTML]{000000} 0.233}                                  & {\color[HTML]{000000} 0.441}           & {\color[HTML]{000000} 15.04}                               & {\color[HTML]{000000} 0.623}                               & {\color[HTML]{000000} 0.269}                                  & {\color[HTML]{000000} 0.425}           \\
{\color[HTML]{000000} UWCNN}                    & {\color[HTML]{000000} 18.70}                               & {\color[HTML]{000000} 0.841}                               & {\color[HTML]{000000} 0.147}                                  & {\color[HTML]{000000} 0.373}           & {\color[HTML]{000000} 22.98}                               & {\color[HTML]{000000} 0.814}                               & {\color[HTML]{000000} 0.203}                                  & {\color[HTML]{000000} 0.388}           \\
{\color[HTML]{000000} TOPAL}                    & {\color[HTML]{000000} 20.86}                               & {\color[HTML]{000000} 0.872}                               & {\color[HTML]{000000} 0.109}                                  & {\color[HTML]{000000} 0.410}           & {\color[HTML]{000000} 20.01}                               & {\color[HTML]{000000} 0.801}                               & {\color[HTML]{000000} 0.214}                                  & {\color[HTML]{000000} 0.409}           \\
{\color[HTML]{000000} UGAN}                     & {\color[HTML]{000000} 21.16}                               & {\color[HTML]{000000} 0.787}                               & {\color[HTML]{000000} 0.149}                                  & {\color[HTML]{000000} 0.438}           & {\color[HTML]{000000} 21.58}                               & {\color[HTML]{000000} 0.787}                               & {\color[HTML]{000000} 0.201}                                  & {\color[HTML]{000000} 0.424}           \\
{\color[HTML]{000000} FunieGAN}                 & {\color[HTML]{000000} 19.39}                               & {\color[HTML]{000000} 0.802}                               & {\color[HTML]{000000} 0.149}                                  & {\color[HTML]{000000} 0.443}           & {\color[HTML]{000000} 22.06}                               & {\color[HTML]{000000} 0.771}                               & {\color[HTML]{000000} 0.182}                                  & {\color[HTML]{000000} 0.418}           \\
{\color[HTML]{000000} Ucolor}                   & {\color[HTML]{000000} 18.12}                               & {\color[HTML]{000000} 0.753}                               & {\color[HTML]{000000} 0.228}                                  & {\color[HTML]{000000} 0.417}           & {\color[HTML]{000000} 20.67}                               & {\color[HTML]{000000} 0.786}                               & {\color[HTML]{000000} 0.230}                                  & {\color[HTML]{000000} 0.427}           \\
{\color[HTML]{000000} PWRNet}                   & {\color[HTML]{000000} 22.50}                               & {\color[HTML]{000000} 0.901}                               & {\color[HTML]{000000} 0.083}                                  & {\color[HTML]{000000} 0.427}           & {\color[HTML]{000000} 23.44}                               & {\color[HTML]{000000} 0.820}                               & {\color[HTML]{000000} 0.208}                                  & {\color[HTML]{000000} 0.427}           \\
{\color[HTML]{000000} ShallowUW}                & {\color[HTML]{000000} 17.66}                               & {\color[HTML]{000000} 0.660}                               & {\color[HTML]{000000} 0.390}                                  & {\color[HTML]{000000} 0.354}           & {\color[HTML]{000000} 22.55}                               & {\color[HTML]{000000} 0.802}                               & {\color[HTML]{000000} 0.205}                                  & {\color[HTML]{000000} 0.379}           \\ \hline
{\color[HTML]{000000} UNet}                     & {\color[HTML]{000000} 17.93}                               & {\color[HTML]{000000} 0.792}                               & {\color[HTML]{000000} 0.200}                                  & {\color[HTML]{000000} 0.389}           & {\color[HTML]{000000} 22.19}                               & {\color[HTML]{000000} 0.802}                               & {\color[HTML]{000000} 0.212}                                  & {\color[HTML]{000000} 0.384}           \\
{\color[HTML]{000000} W/O SPL}              & {\color[HTML]{000000} 22.12}                               & {\color[HTML]{000000} 0.813}                               & {\color[HTML]{000000} 0.121}                                  & {\color[HTML]{000000} 0.401}           & {\color[HTML]{000000} 22.31}                               & {\color[HTML]{000000} 0.807}                               & {\color[HTML]{000000} 0.171}                                  & {\color[HTML]{000000} 0.393}       \\
{\color[HTML]{000000} MDTA+MSFN}           & {\color[HTML]{000000} 22.61}                               & {\color[HTML]{000000} 0.837}                               & {\color[HTML]{000000} 0.140}                                  & {\color[HTML]{000000} 0.402}           & {\color[HTML]{000000} 24.18}                               & {\color[HTML]{000000} 0.823}                               & {\color[HTML]{000000} 0.151}                                  & {\color[HTML]{000000} 0.411}           \\
{\color[HTML]{000000} NFA+GDFN}             & {\color[HTML]{000000} 22.32}                               & {\color[HTML]{000000} 0.831}                               & {\color[HTML]{000000} 0.144}                                  & {\color[HTML]{000000} 0.439}           & {\color[HTML]{000000} 22.56}                               & {\color[HTML]{000000} 0.740}                               & {\color[HTML]{000000} 0.191}                                  & {\color[HTML]{000000} 0.425}           \\
{\color[HTML]{000000} Supervised}               & {\color[HTML]{000000} 22.72}                               & {\color[HTML]{000000} 0.894}                               & {\color[HTML]{000000} 0.085}                                  & {\color[HTML]{000000} 0.413}           & {\color[HTML]{000000} 24.24}                               & {\color[HTML]{000000} 0.840}                               & {\color[HTML]{000000} 0.155}                                  & {\color[HTML]{000000} 0.411} \\
{\color[HTML]{000000} Ours}                     & {\color[HTML]{000000} \textbf{23.23}}                      & {\color[HTML]{000000} \textbf{0.903}}                      & {\color[HTML]{000000} \textbf{0.077}}                         & {\color[HTML]{000000} \textbf{0.446}}  & {\color[HTML]{000000} \textbf{24.40}}                      & {\color[HTML]{000000} \textbf{0.845}}                      & {\color[HTML]{000000} \textbf{0.129}}                         & {\color[HTML]{000000} \textbf{0.431}}  \\ \hline
\end{tabular}
\label{tab:ref_com}
\end{table*}

%% file: tables/noref_com.tex
% Please add the following required packages to your document preamble:
% \usepackage{multirow}
% \usepackage[table,xcdraw]{xcolor}
% If you use beamer only pass "xcolor=table" option, i.e. \documentclass[xcolor=table]{beamer}
\begin{table*}[ht]
\centering
\caption{Quantitative results of comparisons with the state-of-the-art on no-reference datasets. The upper half of the table is the comparison experiment, and the lower half shows the ablation experiment. The best performance is marked in bold, while the second-best performance is underlined.}
\begin{tabular}{l|cc|cc|cc|cc}
\hline
{\color[HTML]{000000} }                         & \multicolumn{2}{c|}{{\color[HTML]{000000} EUVPUN}}                             & \multicolumn{2}{c|}{{\color[HTML]{000000} U45}}                                & \multicolumn{2}{c|}{{\color[HTML]{000000} RUIE}}                               & \multicolumn{2}{c}{{\color[HTML]{000000} UIEB-60}}                            \\ \cline{2-9} 
\multirow{-2}{*}{{\color[HTML]{000000} Method}} & {\color[HTML]{000000} UIQM$\uparrow$} & {\color[HTML]{000000} UCIQE$\uparrow$} & {\color[HTML]{000000} UIQM$\uparrow$} & {\color[HTML]{000000} UCIQE$\uparrow$} & {\color[HTML]{000000} UIQM$\uparrow$} & {\color[HTML]{000000} UCIQE$\uparrow$} & {\color[HTML]{000000} UIQM$\uparrow$} & {\color[HTML]{000000} UCIQE$\uparrow$} \\ \hline
{\color[HTML]{000000} GDCP}                     & {\color[HTML]{000000} 2.751}          & {\color[HTML]{000000} 0.420}           & {\color[HTML]{000000} 2.347}          & {\color[HTML]{000000} 0.415}           & {\color[HTML]{000000} 2.706}          & {\color[HTML]{000000} 0.338}           & {\color[HTML]{000000} 2.196}          & {\color[HTML]{000000} 0.393}           \\
{\color[HTML]{000000} MMLE}                     & {\color[HTML]{000000} 2.737}          & {\color[HTML]{000000} 0.419}           & {\color[HTML]{000000} 2.599}          & {\color[HTML]{000000} 0.423}           & {\color[HTML]{000000} 2.805}          & {\color[HTML]{000000} 0.345}           & {\color[HTML]{000000} 2.197}          & {\color[HTML]{000000} 0.395}           \\
{\color[HTML]{000000} UWCNN}                    & {\color[HTML]{000000} 2.812}          & {\color[HTML]{000000} 0.374}           & {\color[HTML]{000000} 3.151}          & {\color[HTML]{000000} 0.380}           & {\color[HTML]{000000} 2.580}          & {\color[HTML]{000000} 0.245}           & {\color[HTML]{000000} 2.427}          & {\color[HTML]{000000} 0.338}           \\
{\color[HTML]{000000} TOPAL}                    & {\color[HTML]{000000} 2.812}          & {\color[HTML]{000000} 0.414}           & {\color[HTML]{000000} 3.165}          & {\color[HTML]{000000} 0.392}           & {\color[HTML]{000000} 2.871}          & {\color[HTML]{000000} 0.287}           & {\color[HTML]{000000} 2.728}          & {\color[HTML]{000000} 0.373}           \\
{\color[HTML]{000000} UGAN}                     & {\color[HTML]{000000} 2.807}          & {\color[HTML]{000000} 0.438}           & {\color[HTML]{000000} 3.178}          & {\color[HTML]{000000} 0.418}           & {\color[HTML]{000000} 2.882}          & {\color[HTML]{000000} 0.344}           & {\color[HTML]{000000} 2.769}          & {\color[HTML]{000000} 0.394}           \\
{\color[HTML]{000000} FunieGAN}                 & {\color[HTML]{000000} 2.804}          & {\color[HTML]{000000} 0.440}           & {\color[HTML]{000000} 3.202}          & {\color[HTML]{000000} 0.417}           & {\color[HTML]{000000} 2.802}          & {\color[HTML]{000000} \textbf{0.361}}  & {\color[HTML]{000000} \textbf{2.879}} & {\color[HTML]{000000} 0.401}           \\
{\color[HTML]{000000} Ucolor}                   & {\color[HTML]{000000} 2.824}          & {\color[HTML]{000000} 0.430}           & {\color[HTML]{000000} 3.042}          & {\color[HTML]{000000} 0.421}           & {\color[HTML]{000000} 2.607}          & {\color[HTML]{000000} 0.275}           & {\color[HTML]{000000} 2.424}          & {\color[HTML]{000000} 0.395}           \\
{\color[HTML]{000000} PWRNet}                   & {\color[HTML]{000000} 2.806}          & {\color[HTML]{000000} 0.418}           & {\color[HTML]{000000} {3.222}} & {\color[HTML]{000000} 0.423}           & {\color[HTML]{000000} 2.822}          & {\color[HTML]{000000} 0.338}           & {\color[HTML]{000000} 2.700}          & {\color[HTML]{000000} 0.388}           \\
{\color[HTML]{000000} ShallowUW}                & {\color[HTML]{000000} 2.824}          & {\color[HTML]{000000} 0.362}           & {\color[HTML]{000000} 2.980}          & {\color[HTML]{000000} 0.348}           & {\color[HTML]{000000} 2.355}          & {\color[HTML]{000000} 0.222}           & {\color[HTML]{000000} 2.379}          & {\color[HTML]{000000} 0.313}           \\ \hline
{\color[HTML]{000000} UNet}                     & {\color[HTML]{000000} 2.485}          & {\color[HTML]{000000} 0.395}           & {\color[HTML]{000000} 2.936}          & {\color[HTML]{000000} 0.378}           & {\color[HTML]{000000} 2.481}          & {\color[HTML]{000000} 0.247}           & {\color[HTML]{000000} 2.406}          & {\color[HTML]{000000} 0.351}           \\
{\color[HTML]{000000} W/O SPL}              & {\color[HTML]{000000} 2.718}          & {\color[HTML]{000000}  0.417}           & {\color[HTML]{000000}  3.074}          & {\color[HTML]{000000} 0.403}           & {\color[HTML]{000000} 2.757}          & {\color[HTML]{000000} 0.329}           & {\color[HTML]{000000} 2.726}          & {\color[HTML]{000000} 0.367}           \\
{\color[HTML]{000000} MDTA+MSFN}           & {\color[HTML]{000000} 2.723}          & {\color[HTML]{000000} 0.408}           & {\color[HTML]{000000} 3.101}          & {\color[HTML]{000000} 0.401}           & {\color[HTML]{000000} 2.686}          & {\color[HTML]{000000} 0.315}           & {\color[HTML]{000000} 2.711}          & {\color[HTML]{000000} 0.361}           \\
{\color[HTML]{000000} NFA+GDFN}             & {\color[HTML]{000000} 2.710}          & {\color[HTML]{000000} 0.403}           & {\color[HTML]{000000} 3.043}          & {\color[HTML]{000000} 0.396}           & {\color[HTML]{000000} 2.711}          & {\color[HTML]{000000} 0.312}           & {\color[HTML]{000000} 2.713}          & {\color[HTML]{000000} 0.391}           \\
{\color[HTML]{000000} Supervised}               & {\color[HTML]{000000} 2.658}          & {\color[HTML]{000000} 0.419}           & {\color[HTML]{000000} 3.136}          & {\color[HTML]{000000} 0.403}           & {\color[HTML]{000000} 2.781}          & {\color[HTML]{000000} 0.314}           & {\color[HTML]{000000} 2.718}          & {\color[HTML]{000000} 0.377}           \\
{\color[HTML]{000000} Ours}                     & {\color[HTML]{000000} \textbf{2.831}} & {\color[HTML]{000000} \textbf{0.442}}  & {\color[HTML]{000000} \textbf{3.227}} & {\color[HTML]{000000} \textbf{0.440}}  & {\color[HTML]{000000} \textbf{2.891}} & {\color[HTML]{000000} \underline{0.347}}  & {\color[HTML]{000000} \underline{2.789}} & {\color[HTML]{000000} \textbf{0.416}}  \\ \hline
\end{tabular}
\label{tab:noref_com}
\end{table*}

%% file: figtex/results_noref.tex
\begin{figure*}[!ht] 
    \begin{minipage}[b]{1.0\linewidth}
        \begin{minipage}[b]{0.115\linewidth}
            \centering
            \centerline{\includegraphics[width=\linewidth]{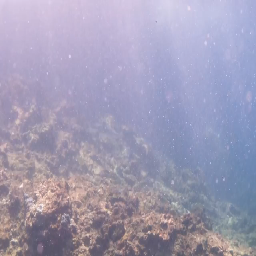}}
        \end{minipage}
        \hfill
        \begin{minipage}[b]{0.115\linewidth}
            \centering
            \centerline{\includegraphics[width=\linewidth]{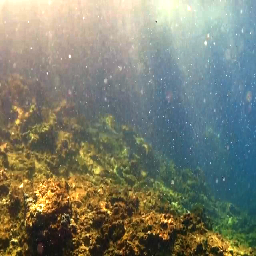}}
        \end{minipage} 
        \hfill
        \begin{minipage}[b]{0.115\linewidth}
            \centering
            \centerline{\includegraphics[width=\linewidth]{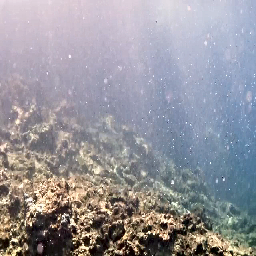}}
        \end{minipage}  
        \hfill
        \begin{minipage}[b]{0.115\linewidth}
            \centering
            \centerline{\includegraphics[width=\linewidth]{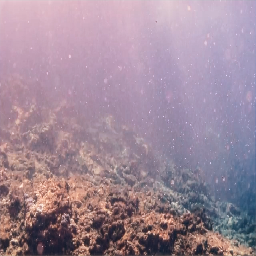}}
        \end{minipage}  
        \hfill
        \begin{minipage}[b]{0.115\linewidth}
            \centering
            \centerline{\includegraphics[width=\linewidth]{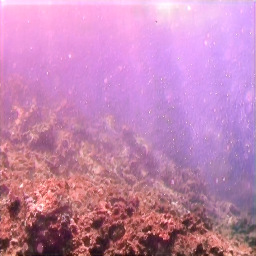}}
        \end{minipage}
        \hfill
        \begin{minipage}[b]{0.115\linewidth}
            \centering
            \centerline{\includegraphics[width=\linewidth]{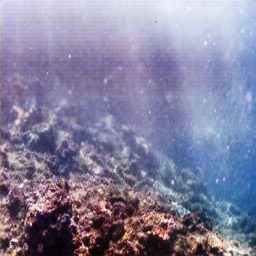}}
        \end{minipage}    
        \hfill
        \begin{minipage}[b]{0.115\linewidth}
            \centering
            \centerline{\includegraphics[width=\linewidth]{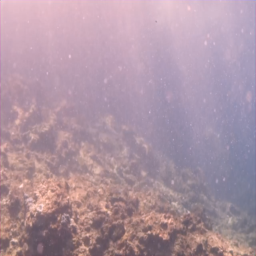}}
        \end{minipage}
        \hfill
        \begin{minipage}[b]{0.115\linewidth}
            \centering
            \centerline{\includegraphics[width=\linewidth]{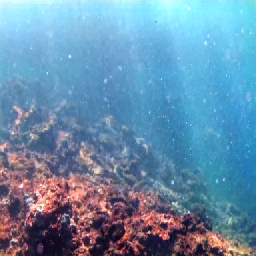}}
        \end{minipage}
    \end{minipage}

    % \begin{minipage}[b]{1.0\linewidth}
    %     \begin{minipage}[b]{0.115\linewidth}
    %         \centering
    %         \centerline{\includegraphics[width=\linewidth]{figs/U45/24input_resized.png}}
    %     \end{minipage}
    %     \hfill
    %     \begin{minipage}[b]{0.115\linewidth}
    %         \centering
    %         \centerline{\includegraphics[width=\linewidth]{figs/U45/24GDCP_resized.png}}
    %     \end{minipage} 
    %     \hfill
    %     \begin{minipage}[b]{0.115\linewidth}
    %         \centering
    %         \centerline{\includegraphics[width=\linewidth]{figs/U45/24MMLE_resized.png}}
    %     \end{minipage}  
    %     \hfill
    %     \begin{minipage}[b]{0.115\linewidth}
    %         \centering
    %         \centerline{\includegraphics[width=\linewidth]{figs/U45/24UWCNN_resized.png}}
    %     \end{minipage}  
    %     \hfill
    %     \begin{minipage}[b]{0.115\linewidth}
    %         \centering
    %         \centerline{\includegraphics[width=\linewidth]{figs/U45/24Ucolor_resized.png}}
    %     \end{minipage}
    %     \hfill
    %     \begin{minipage}[b]{0.115\linewidth}
    %         \centering
    %         \centerline{\includegraphics[width=\linewidth]{figs/U45/24FunieGAN_resized.png}}
    %     \end{minipage}    
    %     \hfill
    %     \begin{minipage}[b]{0.115\linewidth}
    %         \centering
    %         \centerline{\includegraphics[width=\linewidth]{figs/U45/24shallow_resized.png}}
    %     \end{minipage}
    %     \hfill
    %     \begin{minipage}[b]{0.115\linewidth}
    %         \centering
    %         \centerline{\includegraphics[width=\linewidth]{figs/U45/24ours_resized.png}}
    %     \end{minipage}
    % \end{minipage}

    \begin{minipage}[b]{1.0\linewidth}
        \begin{minipage}[b]{0.115\linewidth}
            \centering
            \centerline{\includegraphics[width=\linewidth]{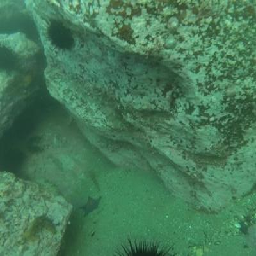}}
        \end{minipage}
        \hfill
        \begin{minipage}[b]{0.115\linewidth}
            \centering
            \centerline{\includegraphics[width=\linewidth]{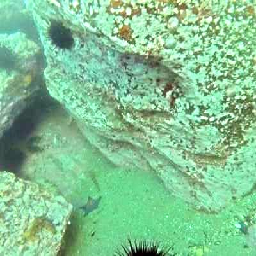}}
        \end{minipage} 
        \hfill
        \begin{minipage}[b]{0.115\linewidth}
            \centering
            \centerline{\includegraphics[width=\linewidth]{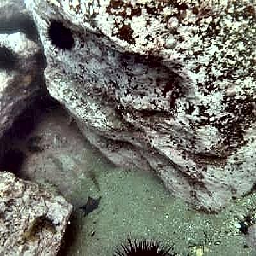}}
        \end{minipage}  
        \hfill
        \begin{minipage}[b]{0.115\linewidth}
            \centering
            \centerline{\includegraphics[width=\linewidth]{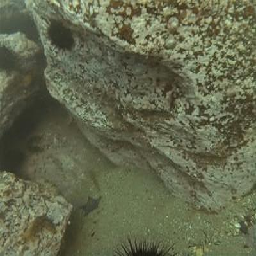}}
        \end{minipage}  
        \hfill
        \begin{minipage}[b]{0.115\linewidth}
            \centering
            \centerline{\includegraphics[width=\linewidth]{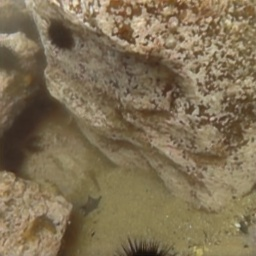}}
        \end{minipage}
        \hfill
        \begin{minipage}[b]{0.115\linewidth}
            \centering
            \centerline{\includegraphics[width=\linewidth]{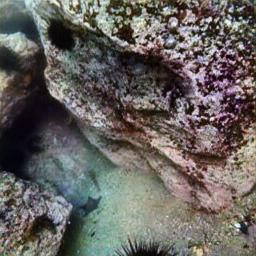}}
        \end{minipage}    
        \hfill
        \begin{minipage}[b]{0.115\linewidth}
            \centering
            \centerline{\includegraphics[width=\linewidth]{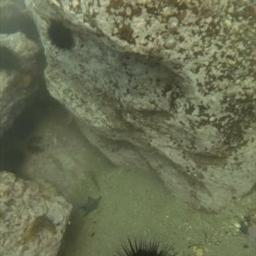}}
        \end{minipage}
        \hfill
        \begin{minipage}[b]{0.115\linewidth}
            \centering
            \centerline{\includegraphics[width=\linewidth]{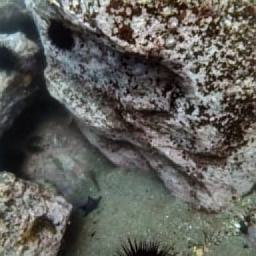}}
        \end{minipage}
    \end{minipage}

    \begin{minipage}[b]{1.0\linewidth}
        \begin{minipage}[b]{0.115\linewidth}
            \centering
            \centerline{\includegraphics[width=\linewidth]{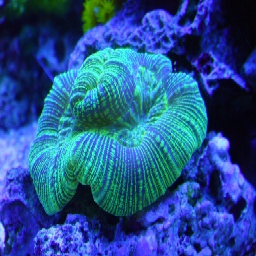}}
            \centerline{(a) Input}\medskip
        \end{minipage}   
        \hfill
        \begin{minipage}[b]{0.115\linewidth}
            \centering
            \centerline{\includegraphics[width=\linewidth]{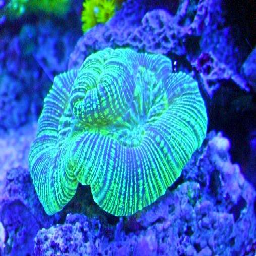}}
             \centerline{(b) GDCP}\medskip
        \end{minipage}  
        \hfill
        \begin{minipage}[b]{0.115\linewidth}
            \centering
            \centerline{\includegraphics[width=\linewidth]{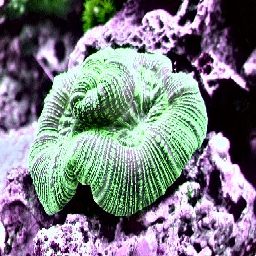}}
            \centerline{(c) MMLE}\medskip
        \end{minipage}   
        \hfill
        \begin{minipage}[b]{0.115\linewidth}
            \centering
            \centerline{\includegraphics[width=\linewidth]{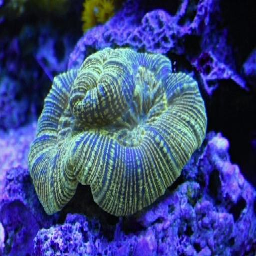}}
            \centerline{(d) UWCNN}\medskip
        \end{minipage}
        \hfill
        \begin{minipage}[b]{0.115\linewidth}
            \centering
            \centerline{\includegraphics[width=\linewidth]{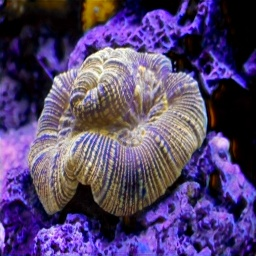}}
            \centerline{(e) Ucolor}\medskip
        \end{minipage}
        \hfill
        \begin{minipage}[b]{0.115\linewidth}
            \centering
            \centerline{\includegraphics[width=\linewidth]{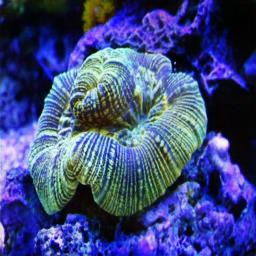}}
            \centerline{(f) FunieGAN}\medskip
        \end{minipage}
        \hfill
        \begin{minipage}[b]{0.115\linewidth}
            \centering
            \centerline{\includegraphics[width=\linewidth]{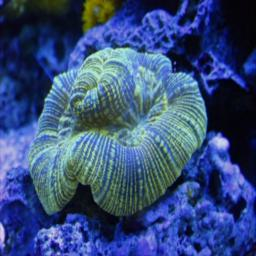}}
             \centerline{(g) ShallowUW}\medskip
        \end{minipage}
        \hfill
        \begin{minipage}[b]{0.115\linewidth}
            \centering
            \centerline{\includegraphics[width=\linewidth]{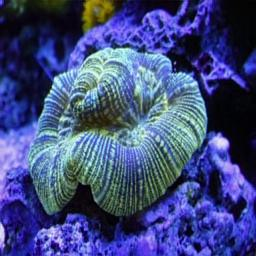}}
            \centerline{(h) Ours}\medskip
        \end{minipage}
    \end{minipage}
    \caption{
    % Visual comparison of different methods for image enhancement on the no-reference datasets. It can be seen that (b), (c) and (e) shows distinct degrees of over-optimization, while (d) and (g) do not fully optimize the images. (f) performs relatively well, however some optimization defects such as ambiguous optimization in the first row and the purple blotches in the third row can still be observed. Similar to full-reference results,  our method~(h) show pleasant results and perform visually better in color tone and details. From top to bottom, each row respectively originates from UIEB-60, U45, RUIE, and EUVPUN.
    Visual comparison of different image enhancement methods with no-reference datasets. It shows that methods (b), (c), and (e) exhibit considerable over-enhancement, while methods (d) and (g) do not fully enhance the images. Method (f) shows relatively satisfactory results albeit ambiguous optimization, such as ambiguous optimization in the first row and purple blotches in the third row. Our method (h) provides superior visual results in terms of color tone and details. The images in the rows from top to bottom, respectively originate from UIEB-60, U45, RUIE, and EUVPUN.
    }
    \label{fig:noref}
\end{figure*}